\documentclass{article}
\usepackage{spconf,amsmath,graphicx}

\newcommand{\ones}{\bm{1}^Ts}

\usepackage{color, colortbl}
\definecolor{Gray}{gray}{0.94}

\usepackage{float}
\usepackage{xcolor}

\usepackage{wrapfig}
\usepackage{caption}
\usepackage{subcaption}

\usepackage{amsmath}
\usepackage{amssymb}
\usepackage{mathtools}
\usepackage{amsthm}
\usepackage{bbm}
\usepackage{bm}
\usepackage{hyperref}

\theoremstyle{plain}
\newtheorem{theorem}{Theorem}[section]
\newtheorem{proposition}[theorem]{Proposition}
\newtheorem*{proposition*}{Proposition}
\newtheorem{lemma}[theorem]{Lemma}

\theoremstyle{definition}

\theoremstyle{remark}

\usepackage{caption}
\title{FW-Shapley: Real-time Estimation of Weighted Shapley Values}
\twoauthors
 {Pranoy Panda \sthanks{both authors contributed equally} \sthanks{This work was done while the author was
a full-time Masters student at IIT Hyderabad}}
{Fujitsu Research India}
 {Siddharth Tandon$^{*}$, Vineeth N Balasubramanian}
{Indian Institute of Technology, Hyderabad}

\begin{document}
\ninept
\maketitle
\begin{abstract}
Fair credit assignment is essential in various machine learning (ML) applications, and Shapley values have emerged as a valuable tool for this purpose. However, in  critical ML applications such as data valuation and feature attribution, the uniform weighting of Shapley values across subset cardinalities leads to unintuitive credit assignments. To address this, weighted Shapley values were proposed as a generalization, allowing different weights for subsets with different cardinalities. Despite their advantages, similar to Shapley values, Weighted Shapley values suffer from exponential compute costs, making them impractical for high-dimensional datasets. 
To tackle this issue, we present two key contributions. Firstly, we provide a weighted least squares characterization of weighted Shapley values. Next, using this characterization, we propose Fast Weighted Shapley (FW-Shapley), an amortized framework for efficiently computing weighted Shapley values using a learned estimator.  We further show that our estimator's training procedure is theoretically valid even though we do not use ground truth Weighted Shapley values during training. 
On the feature attribution task, we outperform the learned estimator FastSHAP by $27\%$ (on average) in terms of Inclusion AUC. For data valuation, we are much faster (14 times) while being comparable to the state-of-the-art KNN Shapley. 
% Through empirical evaluation on two benchmark image datasets, CIFAR10 and SVHN, we demonstrate that FW-Shapley outperforms FastSHAP by $27\%$ (on average) in terms of Inclusion AUC. 
\end{abstract}
\noindent 
%
% \vspace{-10pt}
\section{Introduction}
% \vspace{-10pt}
\textit{Fair} credit assignment is critical in various machine learning (ML) applications. In this context, Shapley values \cite{shapley1953value} have emerged as an important tool. Thus, Shapley values have been widely used for tasks such as interpreting model predictions \cite{lundberg2017unified,sundararajan2020shapley,covert2021improving,ghorbani2020neuron}, data valuation \cite{kwon2021efficient,deutch2021explanations}, model valuation in ensembles \cite{rozemberczki2021shapley}, and feature selection \cite{guha2021cga,patel2021game}.
Shapley values are used for credit assignment because they provide a principled and mathematically grounded approach to allocating credit among the entities involved (e.g., features in the feature attribution task and training data points in the data valuation task). Specifically, they satisfy several key properties that make them well-suited for \textit{fair} credit assignments. These properties include \textit{efficiency}, \textit{symmetry}, \textit{linearity}, and \textit{null player} property. \textit{Efficiency} ensures that the total credit assigned sums up to the overall outcome. \textit{Symmetry} implies that if two entities have the same contribution to all possible coalitions, they should receive equal credit. \textit{Linearity} ensures that the credit assignment is proportional to the contribution of each entity, allowing for consistent and fair credit allocation. Finally, the \textit{null player} property ensures that an entity with no contribution receives zero credit. By considering these properties and the cooperative interactions among the entities, Shapley values enable the fair distribution of credit among features, data points, or models based on their contributions. 

From the viewpoint of game theory, all the Shapley value properties are desirable. However, some properties seem less natural for ML applications leading to unintuitive credit assignments \cite{kwon2021beta,kwon2021efficient}. In particular, Shapley values assign uniform weight to all subset cardinalities while combining their marginal contributions for final credit assignment. This is suboptimal for tasks such as feature attribution and data valuation, as shown by \cite{kwon2022weightedshap} and \cite{kwon2021beta}, respectively. To overcome this limitation, weighted Shapley values were proposed as a generalization of Shapley values \cite{kwon2021beta,kwon2022weightedshap}. Weighted Shapley values (a.k.a. Beta-Shapley values) retain the desirable properties of symmetry, linearity, and the null player property while allowing different weights to be assigned to subsets with different cardinalities. This flexibility in weight assignment enables more meaningful credit assignments in tasks such as feature and data valuation \cite{kwon2022weightedshap,kwon2021beta}.

Like Shapley values, weighted Shapley values also incur an exponential compute cost, thus making its application infeasible for real-world high-dimensional datasets. Sampling-based techniques \cite{kwon2021beta,kwon2022weightedshap} have been used to reduce the compute time, but they must tradeoff runtime with the accuracy of the approximation. 
% Learned estimators such as FastSHAP \cite{jethani2021fastshap} exist to compute shapley values efficiently without dealing with the runtime-accuracy tradeoff. However, it cannot be used for computing weighted Shapley values because it is derived from the weighted least squares characterization of Shapley values \cite{charnes1988extremal}.
In this work, we address this issue by learning an estimator for computing weighted Shapley values. We show results on two tasks: (i) Data valuation -  we are the first to propose a learned estimator. (ii) Feature attribution - here, our work can be seen as a generalization of FastSHAP \cite{jethani2021fastshap} as our framework subsumes it. 

Although a learned estimator reduces compute time without compromising accuracy, training it is non-trivial for two reasons: (i) While Shapley values have a weighted least squares characterization in the literature, a similar optimization-based formulation is absent for weighted Shapley values. (ii) Obtaining a dataset with ground truth weighted Shapley values for high dimensional datasets is infeasible. Thus, an unsupervised objective is necessary for training. Furthermore, optimizing the unsupervised objective should reduce the true estimation error. We propose an objective which satisfies both these constraints. Below we summarize our contributions:
% \vspace{-10pt}
\begin{enumerate}
    \item We introduce an amortized framework, FW-Shapley, for weighted Shapley value computation in real-time via a learned estimator. For this, we derived the weighted least squares characterization of weighted Shapley values and used it to formulate the objective function for our estimator.
    % \vspace{-8pt}
    \item We theoretically validate our estimator's objective function by showing that minimizing the objective is equivalent to minimization of the estimation error.
    % \vspace{-8pt}
    \item We quantitatively show the effectiveness of FW-Shapley across both data valuation and feature attribution tasks on three popular image datasets: CIFAR10, SVHN \& FMNIST.
\end{enumerate}

% \vspace{-8pt}
% Next, we present our methodology, followed by results. We include the related works section in Appendix \ref{sec:rel_works}.
% \vspace{-10pt}
\section{FW-Shapley: Methodology}
% \vspace{-4pt}
\subsection{Notations and Preliminaries}
% \vspace{-5pt}
% \begin{table}[!h]
%     \label{table:notations}
%     \caption{Main notations and their descriptions}
%     \centering
%     \begin{tabular}{|c|p{0.6\linewidth}|}
%         \hline
%         \textbf{Symbol} & \textbf{Description} \\
%         \hline
%         $\bm{x}$, $\mathcal{X}$ & input variable, Domain of input variables \\
%         % \hline
%         % $\mathcal{X}$ & Set containing $x \in \mathbb{R}^d$ \\
%         \hline
%         $\bm{y}$, $\mathcal{Y}$ & Output variable and Domain of output variable for a classification task \\
%         \hline
%         % $\mathcal{Y}$ & Set of possible values for $\bm{y}$, i.e., $\{1, \ldots, K\}$ \\
%         % \hline
%         $d$, $K$ & Number of input features, Number of classes for the classification task\\
%         \hline
%         $p(\bm{x}, \bm{y})$ & Data distribution \\
%         \hline
%         $\bm{s}$ & Binary vector of length $d$ \\
%         \hline
%         $\bm{x}_s$ & Subset of $\bm{x}$ s.t. $\bm{x}_s \coloneqq \{\bm{x_i}\}_{i:s_i=1}$  \\
%         \hline
%         $\bm{1}$, $\bm{0}$ & Vector of ones and zeros of length $d$ \\
%         \hline
%         $\bm{1}^Ts$ & Number of ones in the subset $s$ \\
%         \hline
%     \end{tabular}
% \end{table}
% This paper focuses on applying Weighted Shapley values for data valuation and feature attribution tasks. 
This section introduces the notations used in our paper and provides an overview of weighted Shapley values \cite{kwon2021beta,kwon2022weightedshap}. 
We represent random variables using boldfaced letters and instances of those variables via plain font. Throughout this work, we represent our covariates via $\bm{x} \in \mathcal{X} \subset \mathbb{R}^d$. 
% In the context of data valuation, $d$ refers to the number of training samples and input feature dimensionality for the feature attribution task. 
The output variable $\bm{y} \in \mathcal{Y} = \{1, \ldots, K\}$ for the classification task. In this work, we deal with cooperative games defined over $n$ players. For feature attribution task, players refer to input dimensions, and in data valuation, they refer to training samples. We use $s \in \{0,1\}^n$ to represent subsets of indices $N = \{1,\ldots,n\}$ and define $\bm{x}_s \coloneqq \{\bm{x}_i\}_{i:s_i=1}$ (subset of input features or subset of training samples depending on the target task). Furthermore, we use $\bm{1},\bm{0} \in \mathbb{R}^n$ to represent vectors of all ones and all zeros, respectively. To represent the cardinality of subset $s$, we use the symbol $\bm{1}^{T}s$. The classifier we aim to understand is $f(\bm{x};\eta)$, where $\eta$ represents model parameters. $f_i(\bm{x};\eta)$ is the probability of the $i$-th class for input $\bm{x}$.

% \vspace{-8pt}
\subsection{Weighted Shapley Values}
% \vspace{-5pt}
Weighted Shapley values, a generalization of the well-known Shapley values \cite{shapley1953value}, have been recently introduced to achieve fair credit assignment in machine learning tasks such as  feature attribution and data valuation \cite{kwon2021beta, kwon2022weightedshap}. Weighted Shapley values differ from traditional Shapley values as they allow the assignment of unequal weights to subsets based on their cardinalities, accommodating the varying importance of subset cardinalities. This flexibility improves downstream performance in data valuation tasks and enhances the fidelity of feature attributions \cite{kwon2021beta,kwon2022weightedshap}.

Formally, for a game defined over $n$ players, a value function $v:2^n \rightarrow \mathbb{R}$ and weights $(\alpha, \beta) \in \mathbb{R}^2$, the credit assigned by weighted shapley value for the $i$-th player is defined as follows:
\vspace{-5pt}
\[
\psi_{i}(v) =\sum _{s \subset N; s_{i} \neq 1} \frac{\tilde{w}^{n}_{\alpha,\beta}(\bm{1}^{T}s)}{n}\frac{1}{{n-1 \choose \bm{1}^{T}s}}(v(s\cup\{i\})-v(s))
\vspace{-3pt}
\]
where $\tilde{w}^{n}_{\alpha,\beta}(\bm{1}^{T}s) = {n-1 \choose \bm{1}^{T}s - 1}w^n_{\alpha,\beta}(\bm{1}^{T}s)$ is the normalized weight of the coalition defined over subset of players $s$. This is defined as 

$w^n_{\alpha,\beta}(\bm{1}^{T}s) = \frac{n \times \text{Beta}(\bm{1}^{T}s+\beta-1,n-\bm{1}^{T}s+\alpha)}{\text{Beta}(\alpha,\beta)}$. 

\textbf{Feasible Set for $\bm{\alpha, \beta}$:} In practice, one of $\alpha$ or $\beta$ is set to $1$ so that either we give more importance to smaller cardinality subsets ($\alpha>1, \beta=1$) or  higher cardinality subsets ($\alpha=1, \beta>1$) while computing the credit assignment $\psi_i$ \cite{kwon2021beta, kwon2022weightedshap}. Furthermore, setting the weights too high results in an extremely skewed distribution over the subset cardinalities, which is not ideal for practical applications. Therefore, we select $(\alpha, \beta)$ values from a predefined set $\Delta = \{(1,16),(1,8),(1,4),(1,2), (2,1), (4,1), (8,1), (16,1)\}$ (following \cite{kwon2021beta, kwon2022weightedshap}). $\Delta$ is henceforth referred to as the \textit{feasible set}.

% \vspace{-8pt}
\subsection{Weighted Least Squares Formulation}
% \vspace{-5pt}
In this section, we show that weighted Shapley values for a game defined by the value function $v$ are the solution to a weighted least squares optimization problem. This finding enables us to formulate a least squares objective function for training a weighted Shapley estimator (see Section \ref{subsec:learn_estimator}). To the best of our knowledge, such characterization did not exist for \textit{Weighted Shapley Values} (A similar characterization for Shapley values was shown by \cite{charnes1988extremal}).\\
Below we state our proposition which formalizes the weighted least squares characterization and include the proof in 
 Appendix.
 % \vspace{-5pt}
\begin{proposition}
\label{prop:wls}
Weighted Shapley values for player $z$ can be computed by solving the following weighted least squares optimization problem ($z$ represents an input feature or a training sample):
% \vspace{-2pt}
\[
\psi(z) = \arg\min_{\psi_{z}} \sum_{s\subset N}q(\ones)(v_{z}(s)-v_{z}(\bm{0}) -s^T\psi_{z})^{2}
\vspace{-6pt}
\]
where $q(\ones) = \frac{(n-1)\Tilde{w}^{n}_{\alpha,\beta}}{{n \choose \ones}\ones(n-\ones)}$ refers to the subset weighing function.
This is equivalent to solving the following expectation minimization problem:
% \vspace{-2pt}
\begin{equation}
    \label{eqn:wls}
   \psi(z) = \arg\min_{\psi_{z}} \mathbb{E}_{p(s)}\left[(v_{z}(s)-v_{z}(\bm{0}) -s^T\psi_{z})^{2}\right] 
\end{equation}
where $p(s) \propto \Tilde{w}^{n}_{\alpha,\beta}(\ones)(\ones-1)!(n-\ones-1)!$. $p(s)$ is obtained by normalizing $q(\ones)$. We assume that $\alpha,\beta$ lie in the feasible set and the number of players $n>>\max(\alpha, \beta)$.  
\end{proposition}
% \vspace{-5pt}
Please note the assumptions made in the above proposition hold in practice. For example, for feature attribution and data valuation tasks, let us consider the CIFAR10 dataset which has 1024 features and 60k training samples. The number of features/training samples ($n$) are much larger than the maximum values of $\alpha$ and $\beta$, i.e., 16.

% \vspace{-10pt}
\subsection{Learning the Estimator}
% \vspace{-5pt}
\label{subsec:learn_estimator}

We leverage the weighted least squares characterization (Equation \ref{eqn:wls}) as an objective function for training a parameterized estimator model. By leveraging this approach, the converged estimator can predict the weighted Shapley values without relying on a dataset containing ground truth weighted Shapley values (a similar strategy was followed in \cite{jethani2021fastshap}). The objective function is defined as follows:
\vspace{-6pt}
\begin{equation}
\mathcal{L}(\theta) = \mathbb{E}_{p(z)}\mathbb{E}_{p(s)}\left[(v_{z}(s)-v_{z}(\bm{0}) -s^T\psi_{z})^{2}\right]
\vspace{-6pt}
\end{equation}
In the above equation, $p(s)$ denotes the subset distribution, which is proportional to $\Tilde{w}^{n}_{\alpha,\beta}(\ones)(\ones-1)!(n-\ones-1)!$ (as stated in Proposition \ref{prop:wls}). Additionally, a constraint on the sum of the Shapley values can be imposed to ensure the uniqueness of the weighted Shapley values (Proposition 3 in \cite{kwon2021beta}). This constraint can be incorporated as a regularizer during the training process.

\textbf{Feature Attribution Setting:} In this setting, we operate on image signals, thus we use a UNet \cite{ronneberger2015u} as our estimator to map pixel values to corresponding attribution values. More details in Sec \ref{sec:exp}.

\textbf{Data Valuation Setting:} In this setting, the estimator has to learn the importance of training data points w.r.t. test samples. This is non-trivial as the number of training points can be very large. Thus, we use a batch-learning approach with an attention-based architecture. Below we describe our estimate's form:
\vspace{-5pt}
\[
q\ =\ f_{\theta _{1}}( z_{train}) \ \ \ \ \ \ \ \ k\ =\ f_{\theta _{2}}( z_{val}) \ \ \ \ \ v\ =\ k 
\]
\vspace{-15pt}
\[
\mathrm{Attention}( q,k,v) =\mathrm{sim}( q,k) \times v
\]
\vspace{-12pt}
\[
\mathrm{estimate} \ =\ f_{\theta _{3}}(\mathrm{Attention}( q,k,v)) \
\]
$z_{train}$ and $z_{val}$ are two sampled subsets of data points from the training set. They are named train and val because we wish to find the importance of the ``train" batch on the ``val" samples. $f_{\theta_i}$ are feedforward neural nets with wts $\theta_i$, and $\mathrm{sim}(q,k)_{i,j}$ captures $i$-th train pts similarity with $j$-th val point if their labels match, else its $0$. The masking process is applied to ensure the estimator learns class relevant information.

Next, we provide theoretical justification for minimizing our proposed objective function towards learning an estimator for weighted Shapley values. The proposition below establishes the relationship between minimizing the loss $L(\theta)$ and minimizing the weighted Shapley estimation error.

% \vspace{-9pt}
\begin{proposition}
Under the constraint that the sum of the weighted Shapley values equals some constant, minimizing the loss $\mathcal{L}(\theta)$ minimizes the weighted Shapley estimation error. That is, the objective value $\mathcal{L}(\theta)$ upper bounds the estimation error as follows:
% \vspace{-10pt}
\[
\mathbb{E}_{p(\bm{z})}[||\psi(\bm{z};\theta)-\psi(v_{\bm{z}})||_{2}\leq \sqrt{\sigma(\mathcal{L}(\theta)-\mathcal{L}^*)}]
% \vspace{-5pt}
\]
where $\mathcal{L}^*$ represents loss incurred by exact weighted Shapley values, \& $\sigma$ is a positive constant dependent on $\alpha,\beta, d$.
\end{proposition}
% \vspace{-10pt}
This proposition shows that minimizing the proposed loss results in reduction of true weighted Shapley estimation error. Please refer to Appendix for the proof.

\begin{figure*}
     \centering
     \begin{subfigure}[b]{0.23\textwidth}
         \centering
         \includegraphics[width=0.9\textwidth]{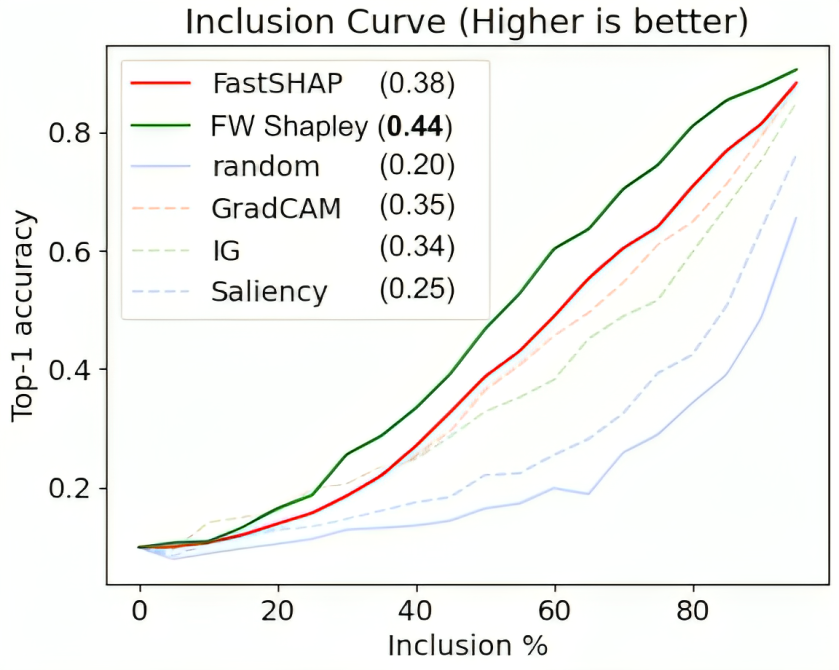}
         \caption{\textbf{Feature Attribution:} Results on CIFAR10 dataset. Inclusion AUC is included in brackets.}
         \label{fig:inc_AUC_cifar}
     \end{subfigure}
     % \hfill
     \hspace{2pt}
     \begin{subfigure}[b]{0.23\textwidth}
         \centering
        \includegraphics[width=0.9\textwidth]{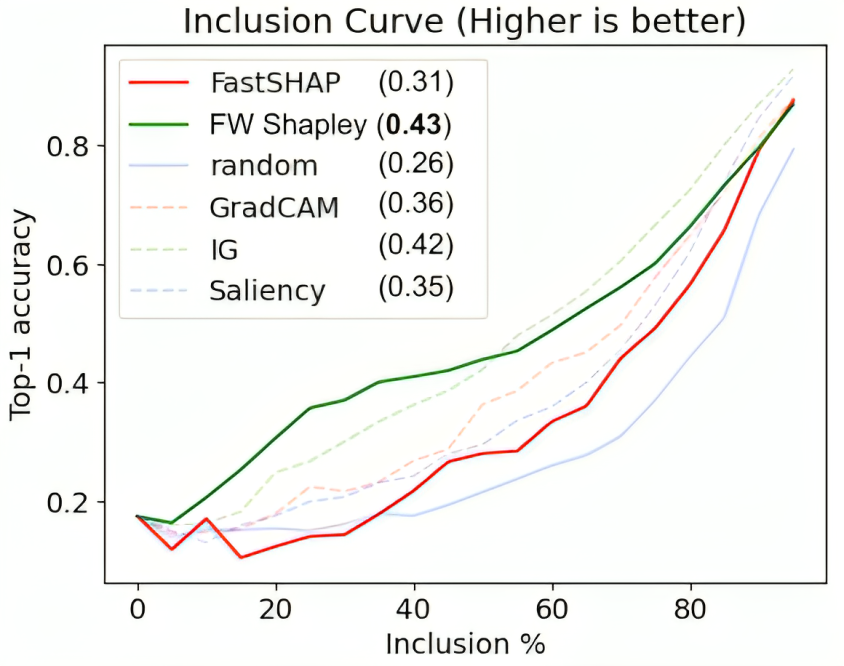}
         \caption{\textbf{Feature Attribution:} Results on SVHN dataset. Inclusion AUC is included in brackets.}
         \label{fig:inc_AUC_svhn}
     \end{subfigure}
     \hfill
     \begin{subfigure}[b]
     {0.23\textwidth}
         \centering
         \includegraphics[width=0.9\textwidth]{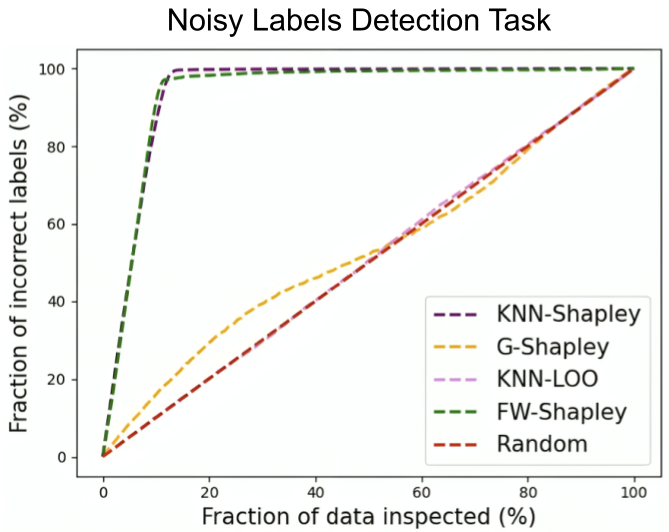}
         \caption{\textbf{Data Valuation:} Results on SVHN dataset. TMC \& Beta-Shapley donot scale to img datasets.}
         \label{fig:noisy_svhn}
     \end{subfigure}
     \hspace{2pt}
     \begin{subfigure}[b]
     {0.23\textwidth}
         \centering
         \includegraphics[width=0.9\textwidth]{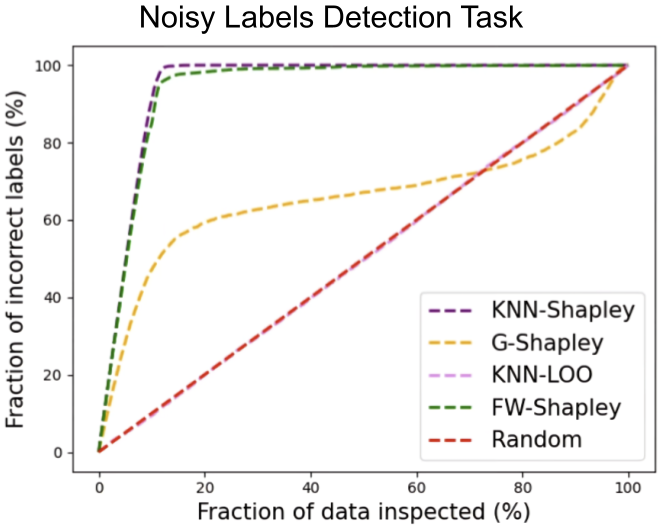}
         \caption{\textbf{Data Valuation:} Results on FMNIST dataset.TMC \& Beta-Shapley donot scale to img datasets.}
         \label{fig:noisy_fmnist}
     \end{subfigure}
        % \caption{Three simple graphs}
        % \label{fig:three graphs}
     % \vspace{-12pt}
\end{figure*}

% \vspace{-10pt}
\subsection{Value Function}
% \vspace{-6pt}
\textbf{Feature Attribution Setting:} Here, a value function computes the classification probability w.r.t. the base model $f(\bm{x};\eta)$, when only a subset of the features $\bm{x_s}$ are observed. To achieve this, we employ a surrogate model \cite{frye2020shapley,jethani2021have}, denoted as $p_{surr}(y|x_s;\zeta)$, which takes a masked feature vector $x_s$ as input. Following \cite{frye2020shapley,jethani2021have}, the parameters $\zeta$ of the surrogate model are learned by minimizing the following loss function:
% \vspace{-6pt}
\[
\mathcal{L}_{surr}(\zeta)=\mathbb{E}_{p(\bm{x})}\mathbb{E}_{p(s)}[D_{KL}(f(\bm{x};\eta) || p_{surr}(y|x_s;\zeta))]
% \vspace{-6pt}
\]
The optimal solution for the above loss corresponds to marginalizing the features from $f(\bm{x};\eta)$ based on their conditional distribution \cite{covert2021explaining}. In other words, $p_{surr}(y|x_s;\zeta^{*}) = \mathbb{E}_{p(\bm{x_s}|y)}[f_y(\bm{x};\eta)|\bm{x_s} = x_s]$. The choice of $p(s)$ does not impact the global optimizer of $\mathcal{L}_{surr}(\zeta)$; therefore, we employ a uniformly random sampler.

While our proposed framework is compatible with any value function, we select this specific value function as it captures the base model's reliance on the \textit{information} conveyed by individual features and feature subsets \cite{frye2020shapley,covert2021explaining}.

\textbf{Data Valuation Setting:} Here, a value function computes performance of the base model had it observed a subset of the training set during training (a counterfactual). To achieve this we use the backbone network (base model without classification head) features and find the K-Nearest Neighbors in the given subset of training data. Then we take a majority vote among the predicted classes of those nearest neighbors. This helps simulate the counterfactual.

% \vspace{-10pt}
\section{Experiments and Results}
\label{sec:exp}
% \vspace{-10pt}
\textbf{Datasets}
To showcase our learned estimator's computational advantages and utility, we conduct experiments on three well-known image datasets - CIFAR10 \cite{krizhevsky2009learning}, SVHN \cite{netzer2011reading} and FMNIST \cite{xiao2017/online}. \\
\textbf{Baselines:}
% In our feature attribution study, we consider FastSHAP as our main baseline since our approach is a generalization of it. Furthermore, we compare our method with three widely used gradient-based feature attribution techniques: GradCAM \cite{selvaraju2017grad}, Saliency \cite{simonyan2013deep}, and IG \cite{pmlr-v70-sundararajan17a}. These gradient-based methods are popular choices for explaining image classifiers due to their efficiency. \\
% For data valuation, we compare against KNN-Shapley \cite{jia2019efficient}, G-Shapley \cite{ghorbani2019data}, KNN-LOO \cite{jia2021scalability} and also a random baseline. We also experimented with Beta-Shapley \cite{kwon2021beta} and TMC Shapley \cite{ghorbani2019data}, however, they are not scalable for high dim datasets such as the ones we experimented on (similar observation in \cite{jia2021scalability}) \\
\underline{Feature attribution task} - a learned estimator already exists in the literature (FastSHAP \cite{jethani2021fastshap} - designed only for this task), which forms our natural baseline. Our objective is to answer these questions: (i) \textit{Can FW-Shapley generate better quality explanations than FastSHAP w.r.t Inclusion AUC (following \cite{jethani2021fastshap})?} (ii) \textit{How does FW-Shapley compare with popular feature attribution methods like GradCAM \cite{selvaraju2017grad}, IG \cite{pmlr-v70-sundararajan17a}, and Saliency \cite{simonyan2013deep}?}. 

\noindent \underline{Data valuation task} - Here, KNN-Shapley(\cite{jia2019efficient}) is the state-of-the-art for fast and accurate Shapley value computation. Thus, we aim to study - \textit{Can FW-Shapley provide data valuations much faster than KNN-Shapley while being as accurate at the same time?}. We also compare with G-Shapley \cite{ghorbani2019data}, KNN-LOO \cite{jia2021scalability}, and the random baseline. We also experimented with Beta-Shapley \cite{kwon2021beta} and TMC Shapley \cite{ghorbani2019data}, however, they are not scalable for high dim datasets such as the ones we experimented on (similar observation in \cite{jia2021scalability}) We use the popular data valuation task of noisy label detection \cite{ghorbani2019data, kwon2021beta,jia2019efficient,jia2021scalability} for our experiments as computing true shapley (weighted) values for high dim datasets is infeasible). 

\noindent \textbf{Implementation details}: We used ResNet-18 networks \cite{he2016deep} as the base classifier $f(\bm{x};\eta)$ and surrogate $p_{surr}(y | \bm{x_s}; \beta)$ for both the learned estimators FW-Shapley and FastSHAP. These networks were pre-trained on ImageNet and fine-tuned on the respective image dataset. We used the official implementation of all the data valuation baselines, and we used \texttt{captum} library \cite{kokhlikyan2020captum} for using feature attribution baselines. Our code is available at \href{https://github.com/sidtandon2014/fw-shapley}{https://github.com/sidtandon2014/fw-shapley}\\
% Following the approach in \cite{jethani2021fastshap}, we configured FW-Shapley and FastSHAP to generate $16 \times 16$ feature attributions for each class. We employed a parameterized UNet model as our estimator, generating feature attributions with an output size of $16 \times 16 \times K$, where $K$ denotes the number of classes.
% For GradCAM, Saliency, and IG, we used the default parameters provided by the \texttt{captum} library \cite{kokhlikyan2020captum}. The official PyTorch implementation of FastSHAP was used for our experiments.
% \vspace{-2pt}
\textbf{Evaluation Metric}
% Assessing the accuracy of Weighted Shapley value estimates necessitates access to ground truth Shapley values, which is computationally infeasible for high-dimensional datasets. Instead, we rely on the inclusion AUC metric that evaluates the effectiveness of an explanation in identifying informative regions within the image. This metric draws upon previous proposals \cite{petsiuk2018rise,hooker2019benchmark,jethani2021have,jethani2021fastshap} and assesses the classification accuracy of the model when pixels are included based on their estimated importance.
% We examine the impact on the model's predictions when we selectively add important features based on each feature attribution method. We conduct this analysis using a set of 1,000 images. We initially assign a label for each image using the original model $f(x; \eta)$ based on the most probable predicted class. We pass each image through $\psi(x,y;\theta)$ to get the feature attributions and use these explanations to produce feature rankings and compute the top-1 accuracy (a measure of agreement with the original model) as we include the most important features, ranging from $0-100$ (masking such features by $0$). 
% This metric aligns with the notion that an effective image explanation should substantially improve the performance of $f(\bm{x};\eta)$ when important features are included \cite{petsiuk2018rise,hooker2019benchmark}.
Obtaining ground-truth Weighted Shapley values is impractical for high-dimensional datasets to assess the accuracy of the estimates. Thus, for feature attribution task, we instead utilize the inclusion AUC metric based on top-1 accuracy, which evaluates the effectiveness of an explanation in identifying informative regions within an image. This metric builds upon prior work \cite{petsiuk2018rise,hooker2019benchmark,jethani2021have,jethani2021fastshap} and measures the model's classification accuracy when pixels are included based on their estimated importance.\\
Similarly, for the data valuation task for noisy label detection, we compute fraction of incorrect labels detected (following \cite{jia2021scalability}).
% To investigate the impact of selectively adding important features, we analyze a set of 1,000 validation images. Initially, we assign labels to the images using the original model $f(\bm{x}; \eta)$ based on the most probable predicted class. Then, we compute feature attributions and use them to compute a ranking over the features and calculate the top-1 accuracy. This accuracy indicates the level of agreement with the original model, as we include increasingly important features ranging from 0 to 100 and mask remaining values by $0$.\\
% The above procedure gives us a curve whose area is termed \textit{Inclusion AUC}. This metric aligns with the notion that an effective image explanation should significantly enhance the performance of original model when important features are included.
\begin{table}[t]
  \centering
  \begin{minipage}{0.47\linewidth}
  \caption{\textbf{Compute Cost (Feature Attribution):} Explanation run-times for 1k images (in secs) on NVIDIA GTX 1080. Here we show results for one dataset as run times are only dependent on input dimensions.}
  % \vspace{-5pt}
  \scalebox{0.9}{
  \begin{tabular}{|c|c|}
    \hline
    \textbf{Datasets} $\rightarrow$ & SVHN \\
    % \vspace{1pt}
    \textbf{Methods} $\downarrow$ &   \\
    \hline
    GradCAM & 17.89 \\
    IG & 147.45 \\
    Saliency & 10.59 \\
    % Random & Data 9 \\
    FastSHAP & \textbf{1.42} \\
    \rowcolor{Gray}
    FW-Shapley (Ours) & \textbf{1.42} \\
    \hline
    \end{tabular}
    \label{table:runtime_feat_attr}
    }
  \end{minipage}
  \hspace{8pt}
  \begin{minipage}{0.47\linewidth}
  \caption{\textbf{Compute Cost (Data Valuation):} Run-times for entire test set of 26k samples (in mins) on NVIDIA GTX 1080. Here runtime depends on size of train and test set.}
  % \vspace{-5pt}
  \scalebox{0.92}{
  \begin{tabular}{|c|c|}
    \hline
    \textbf{Datasets} $\rightarrow$ & SVHN \\
    % \vspace{1pt}
    \textbf{Methods} $\downarrow$ &  \\
    \hline
    KNN-Shapley & $456$ \\
    KNN-LOO & $145$ \\
    G-Shapley & $734$ \\
    % Random & Data 9 \\
    TMC-Shapley & $>800$ \\
    Beta-Shapley & $>800$ \\
    \rowcolor{Gray}
    FW Shapley (Ours) & \textbf{1.3} \\
    \hline
    \end{tabular}
    \label{table:runtime_data_val}
    }
  \end{minipage}
  % \vspace{-15pt}
\end{table}
% \vspace{-13pt}
\subsection{Results}
% \vspace{-5pt}
% Figures (a) and (b) show the inclusion curve for CIFAR10 and SVHN datasets, respectively, these figures also show the inclusion AUC comparisons. We also report the run time comparison with different attribution methods in Table \ref{table:runtime_feat_attr}. 
% Please note that our work is a generalized version of the FastSHAP methodology. Thus overall training and evaluation times remain the same for both. The only difference is that we treat the weights $\alpha, \beta$ used in the subset sampling distribution $p(s)$ as hyperparameters during training.
The results across Figures (a) and (b) show that FW-Shapley outperforms all baseline methods while simultaneously being the fastest feature attribution method w.r.t. explanation time (Table 1). 

Similarly, from Table 2 and Figures (c) and (d) we can infer that in the data valuation setting (on the noisy label detection task) it achieves a \textit{14x speedup} (including training time of 30 mins) w.r.t. the state-of-the-art KNN-Shapley while being equally accurate in detection. Thus, we empirically show that our theoretically grounded solution FW-Shapley is the fastest weighted shapley estimator.

% Figure \ref{fig:inc_training_time_fastshap} has been directly borrowed from FastSHAP \cite{jethani2021fastshap} paper 
% \vspace{-15pt}
\section{Related Work}
% \vspace{-8pt}
\label{sec:rel_works}
Over the past decade, researchers have devoted significant attention to efficiently approximating Shapley values \cite{michalak2013efficient,chen2018shapley,ancona2019explaining,lundberg2020local,covert2021improving}. While some approaches provide fast computation by leveraging model-specific characteristics and feature removal techniques, they often introduce bias. On the other hand, model-agnostic methods offer greater flexibility but face the accuracy-efficiency tradeoff. A recent model-agnostic method called FastSHAP \cite{jethani2021fastshap} has emerged as the fastest (in an amortized sense) Shapley estimator in this category. It focuses on the feature attribution task and learns Shapley values by minimizing the weighted least squares characterization \cite{charnes1988extremal}.

However, Shapley values, by default, assign equal weight to all subset cardinalities, which may not be optimal for certain tasks like feature attribution and data valuation \cite{kwon2021beta,kwon2022weightedshap}. To address this limitation, weighted Shapley values were introduced, aiming to improve the assignment of credits. Unfortunately, existing techniques developed for Shapley values computation cannot be directly applied to compute weighted Shapley values efficiently. To the best of our knowledge, our work represents the first attempt to compute model-agnostic weighted Shapley values efficiently. 

% \vspace{-8pt}
% \section{Conclusion}
% \vspace{-8pt}
% In this work, we developed FW-Shapley, a learning-based approach to compute weighted Shapley values efficiently. Our method allows for more meaningful credit assignments, surpassing the limitations of uniform weighting in FastSHAP. We have demonstrated the computational benefits of our learned estimator on high-dimensional image datasets, CIFAR10 and SVHN. Through experiments, we have shown that FW-Shapley outperforms FastSHAP in generating high-quality explanations, as measured by the Inclusion AUC metric.
% % Additionally, our method has been compared with popular gradient-based feature attribution techniques, further highlighting its effectiveness. 
% Overall, FW-Shapley offers a valuable solution for fair credit assignment which in turn helps provide improved interpretability in machine learning models.
% \vspace{-12pt}
\section{Appendix: Proofs}
\label{sec:appx}
\begin{figure*}[!htp]
    \centering
    \begin{minipage}{0.32\linewidth}
        \centering
        \includegraphics[scale=0.12]{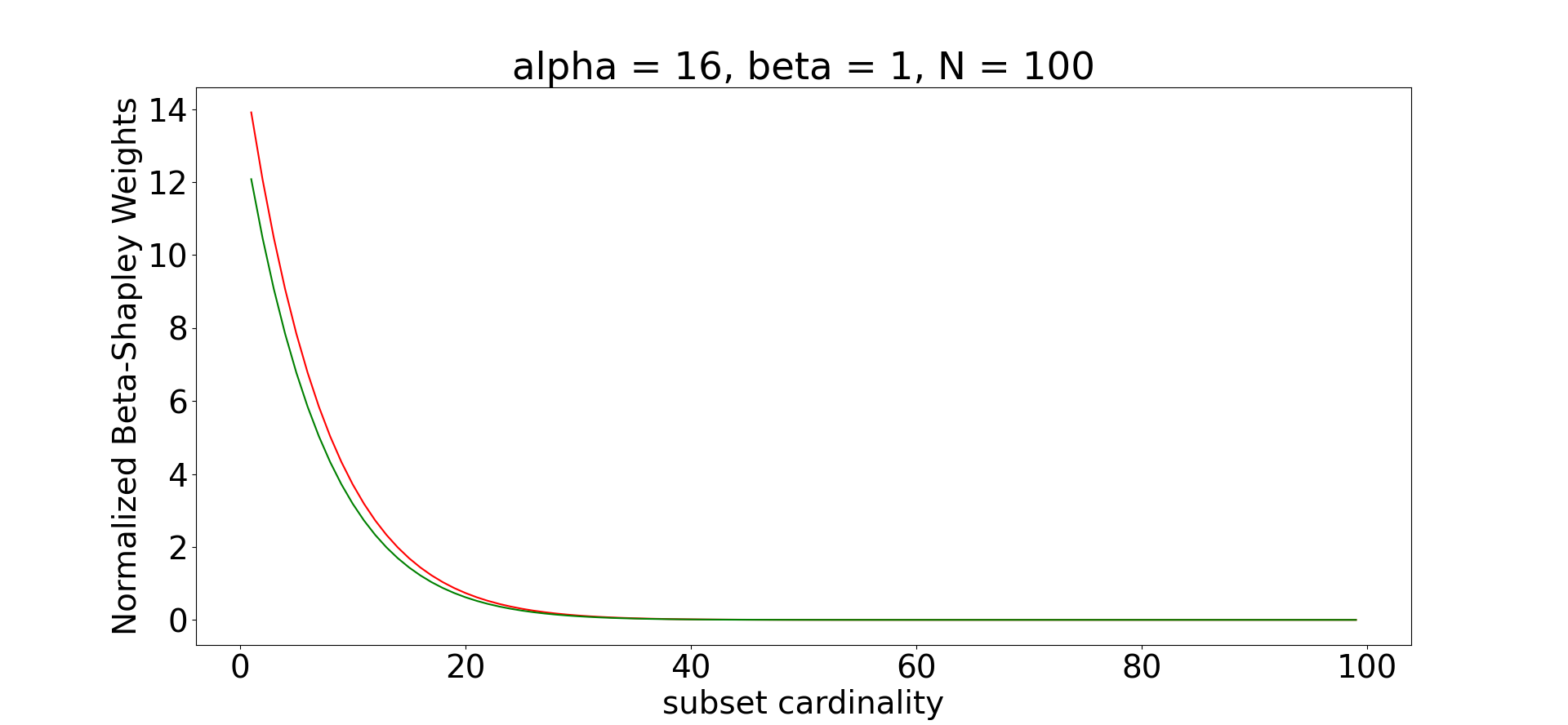}
        \caption*{(a) $n = 100$}
    \end{minipage}
    \hfill
    \begin{minipage}{0.32\linewidth}
        \centering
        \includegraphics[scale=0.12]{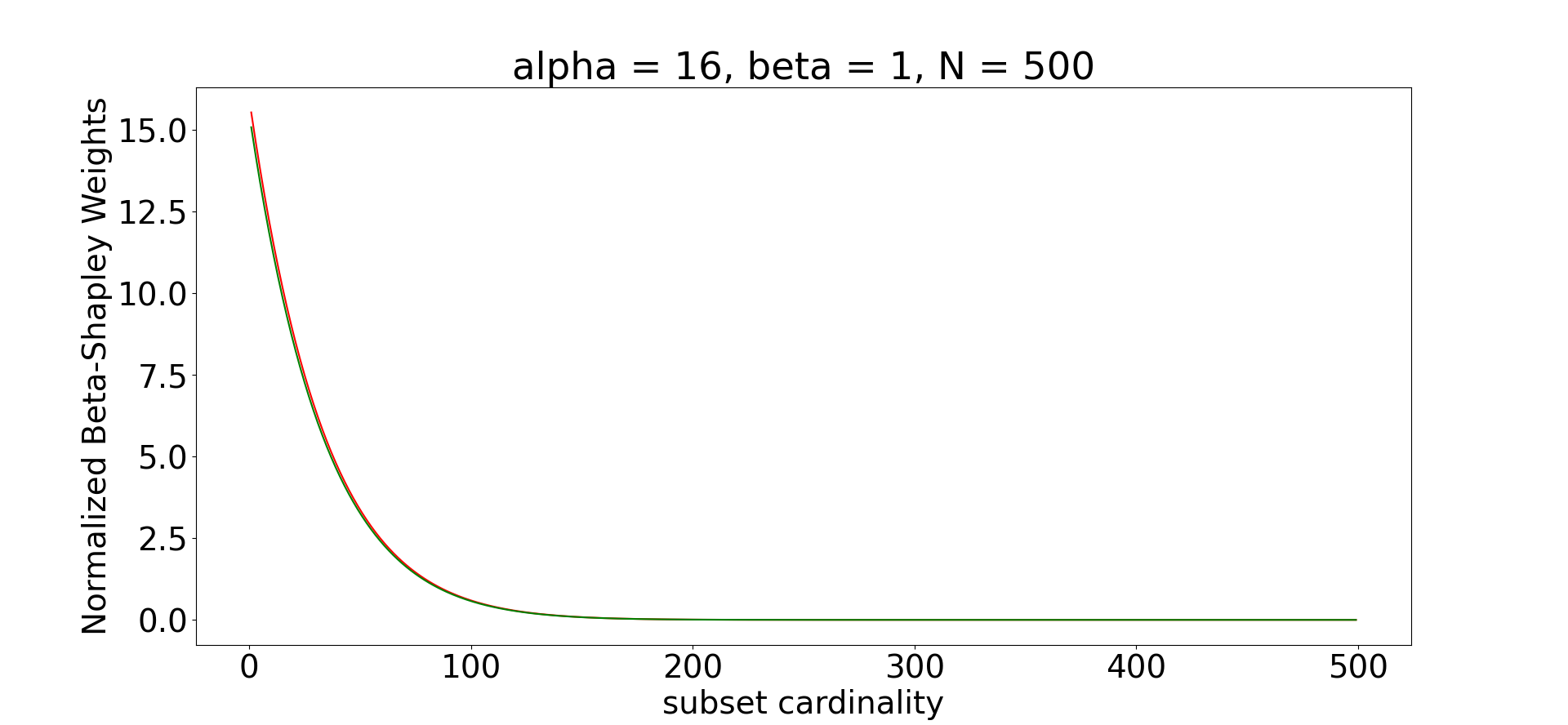}
        \caption*{(b) $n = 500$}
    \end{minipage}
    \hfill
    \begin{minipage}{0.32\linewidth}
        \centering
        \includegraphics[scale=0.12]{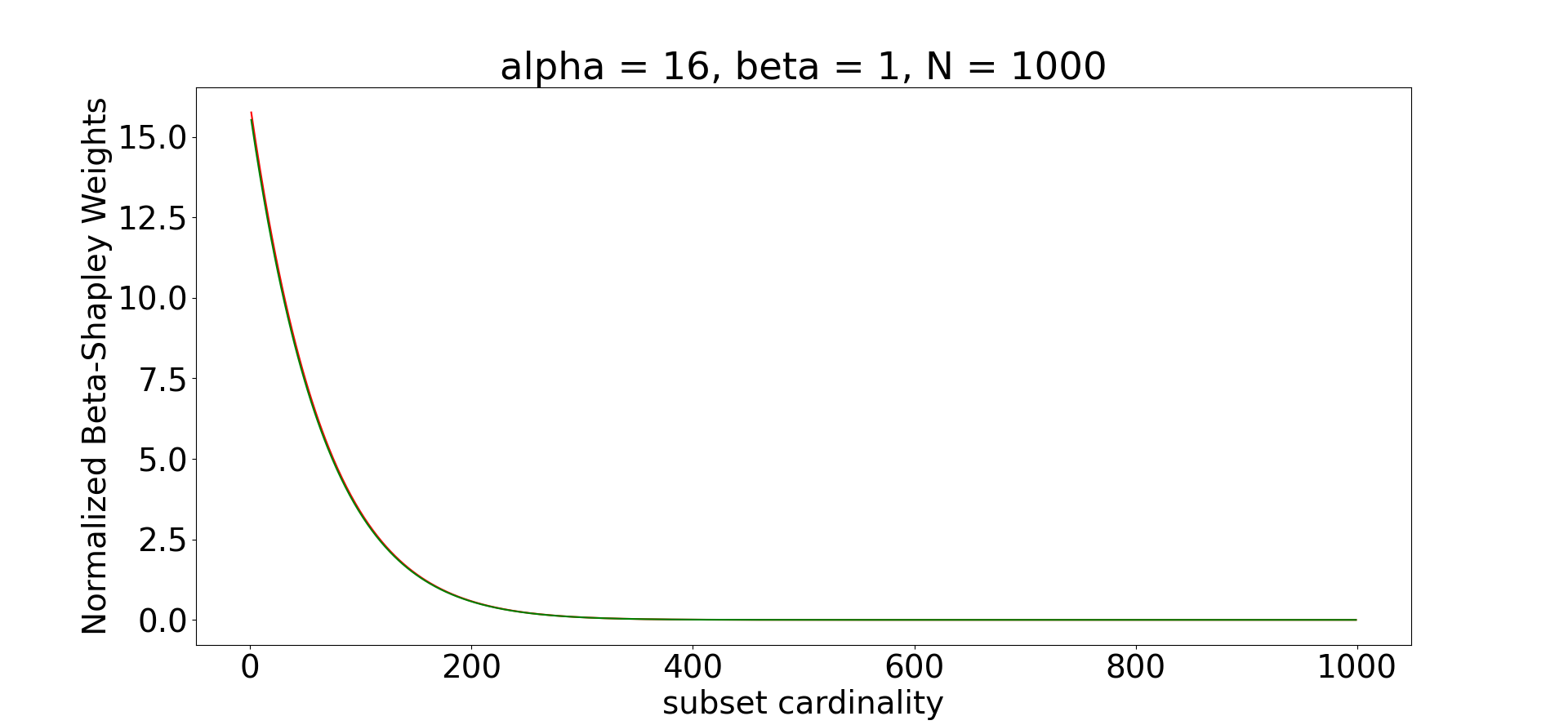}
        \caption*{(c) $n = 1000$}
    \end{minipage}
    
    \caption{\textbf{Numerical Validation of Lemma 5.1} (\textcolor{red}{\textit{Red}} - $\Tilde{w}^{n}_{\alpha,\beta}(k-1)$, \textcolor{green}{\textit{Green}} - $\Tilde{w}^{n}_{\alpha,\beta}(k)$, $\alpha=16, \ \beta=1$) As the number of features/data samples $n$ approaches $500$ (very small in comparison to today's available datasets), the curves for $\Tilde{w}^{n}_{\alpha,\beta}(k)$ and $\Tilde{w}^{n}_{\alpha,\beta}(k-1)$ become indistinguishable.}
    \label{fig:lemma_a16b1}
\end{figure*}

\begin{figure*}[!htp]
    \centering
    \begin{minipage}{0.33\linewidth}
        \centering
        \includegraphics[scale=0.28]{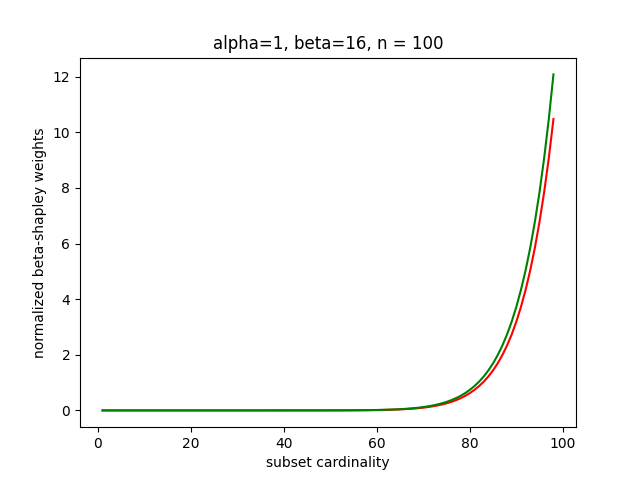}
        \caption*{(a) $n = 100$}
    \end{minipage}
    \hfill
    \begin{minipage}{0.33\linewidth}
        \centering
    \includegraphics[scale=0.28]{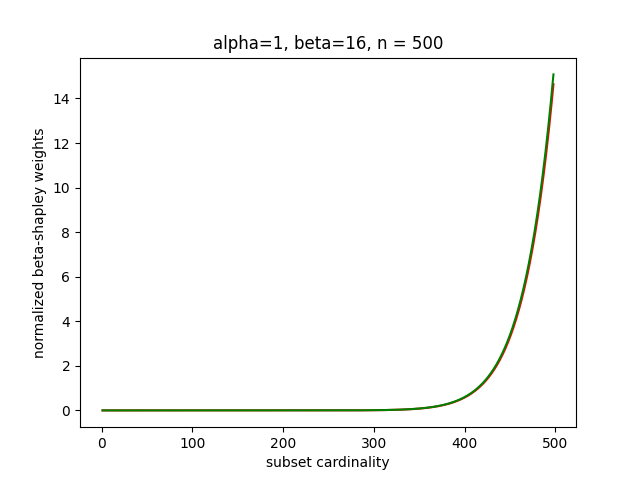}
        \caption*{(b) $n = 500$}
    \end{minipage}
    \hfill
    \begin{minipage}{0.33\linewidth}
        \centering
        \includegraphics[scale=0.28]{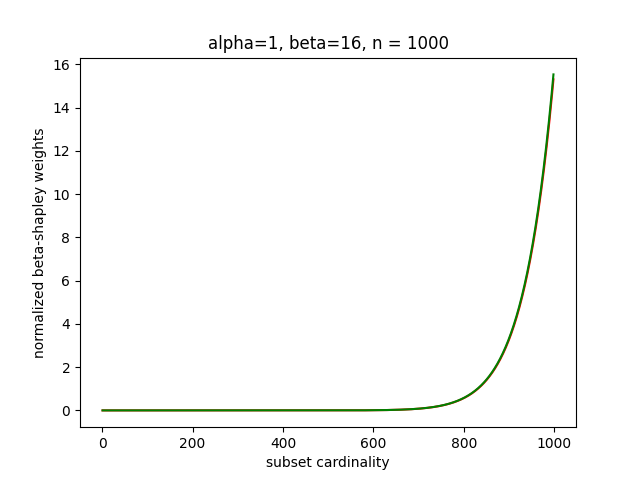}
        \caption*{(c) $n = 1000$}
    \end{minipage}
    
    \caption{\textbf{Numerical Validation of Lemma 5.1} (\textcolor{red}{\textit{Red}} - $\Tilde{w}^{n}_{\alpha,\beta}(k-1)$, \textcolor{green}{\textit{Green}} - $\Tilde{w}^{n}_{\alpha,\beta}(k)$, $\alpha=1, \ \beta=16$) As the number of features/data samples $n$ approaches $500$ (very small in comparison to today's available datasets), the curves for $\Tilde{w}^{n}_{\alpha,\beta}(k)$ and $\Tilde{w}^{n}_{\alpha,\beta}(k-1)$ become indistinguishable.}
    \label{fig:lemma_a1b16}
\end{figure*}

% \vspace{-7pt}
\textbf{Proof for Proposition 2.1}
(The number of features/training samples represents the number of players $n$ in the following proofs.)
% \vspace{-5pt}
\label{sec:prop1_proof}
\begin{lemma}
    \label{lemma1}
    Given that $\alpha,\beta$ lie in the feasible set and the number of players $n>>\max(\alpha, \beta)$, the normalized weight function satisfies the following property: $\Tilde{w}^{n}_{\alpha,\beta}(k-1) \approx \Tilde{w}^{n}_{\alpha,\beta}(k) \ \forall k$
    % \vspace{-5pt}
    \begin{proof}
        To show this holds for the entire feasible set, we analyze the following two cases - (i) $\beta=1$ and $\alpha>1$ (pertinent for data valuation task \cite{kwon2021beta}) (ii) $\alpha=1$ and $\beta>1$ (pertinent for feature attribution task in vision datasets). As a preliminary for this analysis, we first evaluate the ratio between $\Tilde{w}^{n}_{\alpha,\beta}(k)$ and $\Tilde{w}^{n}_{\alpha,\beta}(k-1)$ below. 
% \vspace{-7pt}
\begin{equation*}
\frac{\Tilde{w}^{n}_{\alpha,\beta}(k)}{\Tilde{w}^{n}_{\alpha,\beta}(k-1)} = \frac{{n-1 \choose k-1}}{{n-1 \choose k-2}}\frac{w^{n}_{\alpha,\beta}(k)}{w^{n}_{\alpha,\beta}(k-1)} = \frac{n-k+1}{k-1}\frac{w^{n}_{\alpha,\beta}(k)}{w^{n}_{\alpha,\beta}(k-1)}
% \vspace{-4pt}
\end{equation*}
$w^{n}_{\alpha,\beta}(k) = \frac{n}{\text{Beta}(\alpha,\beta)\Gamma(n+\alpha+\beta-1)}\Gamma(k+\beta-1)\Gamma(n-k+\alpha)$ 

\noindent (From the definition of $w^{n}_{\alpha,\beta}(k)$ in Sec 2.2). We substitute this in the above equation to further reduce it as follows.
% \vspace{-7pt}
\begin{equation}
    \label{eqn:ratio}
    \frac{\Tilde{w}^{n}_{\alpha,\beta}(k)}{\Tilde{w}^{n}_{\alpha,\beta}(k-1)} = \frac{n-k+1}{k-1}\frac{k+\beta-2}{n-k+\alpha} \ \ \ \ (\because \Gamma(n+1) = n\Gamma(n))
    % \vspace{-5pt}
\end{equation}
\textit{Case (i)} : [$\beta=1, \alpha>1$] Substituting the value of $\beta=1$ in Equation \ref{eqn:ratio} and using the prior assumption that $n >> \alpha$, yeilds that $\Tilde{w}^{n}_{\alpha,\beta}(k-1) \approx \Tilde{w}^{n}_{\alpha,\beta}(k) \ \forall k$.
% \vspace{-7pt}
\[
\frac{\Tilde{w}^{n}_{\alpha,\beta}(k)}{\Tilde{w}^{n}_{\alpha,\beta}(k-1)} = \frac{n-k+1}{(n+\alpha-1)-k+1} \approx 1 \ \ \ \ (\because n >> \alpha-1)
\vspace{-7pt}
\]
\textit{Case (ii)} : [$\alpha=1, \beta>1$]
% \vspace{-7pt}
\[
\frac{\Tilde{w}^{n}_{\alpha,\beta}(k)}{\Tilde{w}^{n}_{\alpha,\beta}(k-1)} = \frac{k+\beta-2}{k-1} = \frac{(k+\beta-1) - 1}{k-1}
% \vspace{-7pt}
\]

The value of the above ratio lies in the range $[\frac{n+\beta-2}{n-1},\beta,]$ as $k$ varies from $2$ to $n$. Given that $\beta$ lies in the feasible set and $n>>\beta$, the range of the ratio = $[1,\beta]$. This ratio falls off very fast from $\beta$ to $1$, thus, for all high cardinality subsets ($k>n/2$), ratio is close to $1$.

For subsets with $k < \beta << n$, significantly greater than one, however, $\Tilde{w}^{n}_{\alpha,\beta}(k)$ is very small as we intentionally want to down weight smaller cardinality subsets (given we are in the setting $\alpha=1, \beta>1$). So, although ratio is significant, the weightage given to such subsets while computing the weighted Shapley is negligible. We further numerically validate this lemma in Figure \ref{fig:lemma_a16b1} \& \ref{fig:lemma_a1b16}
\end{proof}
\end{lemma}

% \vspace{-20pt}
\begin{proof}
    \label{proof:WLS}
    \textbf{[Start of main proof]} To show this result, we use the \textit{Generalized Shapley Value Theorem} \cite{charnes1988extremal}. This theorem can be trivially extended for any linear constraint. Following this, we simplify the result and show that plugging our proposed subset weighting function $q(.)$ yields weighted Shapley values (under some realistic assumptions w.r.t. the feature attribution and data valuation settings).
    $\triangleright	$ \textit{Proposed Sampler Substitution}: Here we substitute the proposed sample weighing function $q(\ones) = \frac{(n-1)\Tilde{w}^{n}_{\alpha,\beta}}{{n \choose \ones}\ones(n-\ones)}$ into the extended generalized shapley values. Below is the formula of extended generalized shapley values.
    % \vspace{-10pt}
    \[
        \psi_i = \frac{C}{n} + \underbrace{\frac{1}{\sum_{k=1}^{n-1}{n-2 \choose k-1}q(k)}}_{\textcircled{1}} \sum_{s\subset N; i \in s}\bigg( \underbrace{\frac{n-\ones}{n}q(\ones)}_{\textcircled{2}}v(s)
        % \vspace{-12pt}
        \]
        
    \[- \underbrace{\frac{\ones-1}{n}q(\ones - 1)}_{\textcircled{3}}v(\ones -1 )  \bigg)
    \vspace{-7pt}
    \]
    Now, we analyze the products $\textcircled{1} \times \textcircled{2}$ and $\textcircled{1} \times \textcircled{3}$ 
    % \vspace{-5pt}
    \begin{equation}
        \label{eqn:coef1}
        \textcircled{1} \times \textcircled{2} = \Tilde{w}^{n}_{\alpha,\beta}(k)\frac{n-1}{n-\Tilde{w}^{n}_{\alpha,\beta}(n)} \frac{(k-1)!(n-k)!}{n!}
    \end{equation}
    % \vspace{-7pt}
    \begin{equation}
        \label{eqn:coef2}
        \textcircled{1} \times \textcircled{3} = \Tilde{w}^{n}_{\alpha,\beta}(k-1)\frac{n-1}{n-\Tilde{w}^{n}_{\alpha,\beta}(n)} \frac{(k-1)!(n-k)!}{n!}
    \end{equation}

For practical applicability, $\alpha, \beta$ lie in the feasible set, i.e., $\{(1,16),(1,8),(1,4), (4,1), (8,1), (16,1)\}$ and in this range, $\lim_{n\rightarrow \infty} \Tilde{w}^{n}_{\alpha,\beta}(n) = \max(\alpha,\beta)$. The difference between $\Tilde{w}^{n}_{\alpha,\beta}(n)$ and $\max(\alpha,\beta)$ falls of exponentially with $n$ and even for $n=1000$, the difference is as small as $0.2$. Thus, it is a fair approximation to make. Using this 
observation and the assumption that $n>>\max(\alpha,\beta)$, we set $\frac{n-1}{n-\Tilde{w}^{n}_{\alpha,\beta}(n)} \approx \frac{1-\frac{1}{n}}{1-\frac{\max(\alpha,\beta)}{n}}$ to 1.

Using the above finding and Lemma \ref{lemma1}, we can infer that $\textcircled{1} \times \textcircled{2} = \textcircled{1} \times \textcircled{3} = \Tilde{w}^{n}_{\alpha,\beta}(k)\frac{(k-1)!(n-k)!}{n!}$. Substituting this into ext. gen. shapley values gives us the weighted Shapley form \cite{kwon2021beta}.
\begin{equation*}
    \label{eqn:weighted_shapley_final}
        \psi_i = \sum_{s\subset N; i \in s} \Tilde{w}^{n}_{\alpha,\beta}(\ones)\frac{(\ones-1)!(n-\ones)!}{n!} \bigg( v(s) - v(s_{\backslash i} )  \bigg) + \frac{C}{n}
\end{equation*}
\end{proof}
% \vspace{-22pt}
\subsection{Proof for Proposition 2.2}
% \vspace{-7pt}
\begin{proof}
    First, we show that for a single player $z$, the expected loss for the prediction made by the estimator $\psi(\bm{z};\theta)$ is $\mu$-strongly convex. Then, we use the Lagrangian multiplier to account for the linear constraint on the sum of the weighted Shapley weights and show that it is also $\mu$-strongly convex. Lastly, we use the first-order strong convexity condition to derive an upper bound on the expected error.
    
    \noindent $\triangleright	$ \textit{For a single player $\bm{z}$, the expected loss is $\mu$-strongly convex}
    
    \noindent To prove that the loss is $\mu$-strongly convex, we show that the minimum eigenvalue of the hessian of the loss is positive. See similar results in \cite{covert2022learning,covert2021improving,simon2020projected}
    
    \noindent To compute the hessian, we re-write the loss equation for a single player $z$ and differentiate it twice w.r.t. $\psi$: $
    \nabla^{2}_{\psi} \mathcal{L}_{z}(\psi)  = 2\mathbb{E}_{p(s)}[ss^T]$
    
    Let $A \coloneqq \mathbb{E}_{p(s)}[ss^T]$, where $A_{ii} = P(s_i=1) = \sum_{k=1}^{n}{n-1 \choose k-1}p_k$ and $A_{ij} = P(s_i=s_j=1) = \sum_{k=2}^{n}{n-2 \choose k-2}p_k$
    
    As shown by Simon et al. \cite{simon2020projected} and Covert et al. \cite{covert2022learning}, $A$ has the following form $A=(a-b)I_n + b\bm{1}\bm{1}^T$ where $a = A_{ii}$, $b = A_{ij}$ and the minimum eigenvalue is given by $\lambda_{min} = a-b$. With respect to our subset sampling distribution, $p_k = Q^{-1}\frac{(n-1)\Tilde{w}^{n}_{\alpha,\beta}(k)}{{n \choose k }k(n-k)}$, where $Q=\sum_{j=1}^{n-1}\frac{(n-1)\Tilde{w}^{n}_{\alpha,\beta}(j)}{j(n-j)}$ is the normalization constant obtained by solving $\sum_{k=1}^{n-1}{n \choose k}p_k=1$. Simplifying  $\lambda_{min} = a-b $ gives us the following: $
    \lambda_{min} = \frac{1-\Tilde{w}^{n}_{\alpha,\beta}(n)}{Q}
    $
    . $\lambda_{min}$ is strictly positive as both denominator and numerator are strictly positive.
    
    \noindent Thus, we prove that $\mathcal{L}_{z}(\psi)$ is $\mu$-strongly convex where $\mu = 2\lambda_{min}$.

    % \vspace{5pt}
    
    \noindent $\triangleright	$ \textit{Deriving Upper bound via applying strong convexity condition}
    
    This follows directly from Theorem 1 in \cite{covert2022learning} as we have already shown above that for any data point $z = (x,y)$, the expected weighted shapley loss $\mathcal{L}_z$ is $\mu$-strongly convex. Thus, 
    % \vspace{-5pt}
    \[
    \mathbb{E}_{p(\bm{z})}[||\psi(\bm{z};\theta)-\psi(v_{\bm{z}})||_{2}\leq \sqrt{\sigma(\mathcal{L}(\theta)-\mathcal{L}^*)}]
    \]
    where $\sigma = \frac{2}{\mu} > 0$ as $\mu = \frac{2(1-\Tilde{w}^{n}_{\alpha,\beta}(n))}{Q}>0$
\end{proof}

\clearpage
\scriptsize
\bibliographystyle{IEEEbib}
\bibliography{main}

\end{document}